\documentclass[10pt, a4paper]{article}
\usepackage{lrec-coling2024} 

\usepackage{natbib}
\usepackage{multibib}
\usepackage{graphicx}
\usepackage{tabularx}
\usepackage{soul}
\usepackage{titlesec}
\usepackage{csquotes}

\usepackage{hyperref}
\usepackage{xstring}

\usepackage{color}

\usepackage{multirow}
\usepackage{arydshln}
\usepackage{amsmath,mathtools,amsfonts}
\usepackage{booktabs}
\usepackage{tikz}
\usepackage{enumitem}

\newcommand{\mg}{\textcolor{black}}
\newcommand{\sara}{\textcolor{black}}

\makeatletter
\newcommand{\thickhline}{%
   \noalign {\ifnum 0=`}\fi \hrule height 1pt
   \futurelet \reserved@a \@xhline
}
\newcolumntype{"}{@{\hskip\tabcolsep\vrule width 1pt\hskip\tabcolsep}}
\makeatother

\newcommand\blfootnote[1]{%
  \begingroup
  \renewcommand\thefootnote{}\footnote{#1}%
  \addtocounter{footnote}{-1}%
  \endgroup
}

\title{How do Hyenas deal with Human Speech? \\
Speech Recognition and Translation with ConfHyena}

\name{Marco Gaido\includegraphics[scale=0.1]{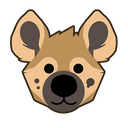}\includegraphics[scale=0.06]{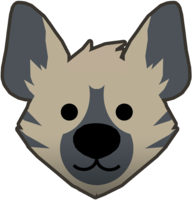}, Sara Papi\includegraphics[scale=0.1]{spotted_hyena96.png}\includegraphics[scale=0.06]{7139-yeenbean.png}, Matteo Negri\includegraphics[scale=0.06]{7139-yeenbean.png}, Luisa Bentivogli\includegraphics[scale=0.06]{7139-yeenbean.png}} 

\address{\includegraphics[scale=0.06]{7139-yeenbean.png} Fondazione Bruno Kessler, Italy \\
\texttt{\{mgaido,spapi,negri,bentivo\}@fbk.eu}\\}

\abstract{
The attention mechanism, a cornerstone of state-of-the-art neural models, faces computational hurdles in processing long sequences due to its quadratic complexity.
Consequently, research efforts in the last few years focused on finding more efficient alternatives. 
Among them, Hyena \citep{Poli2023HyenaHT} stands out for achieving competitive results in both language modeling and image classification, while offering sub-quadratic memory and computational complexity.
Building on these promising results, we propose ConfHyena, a Conformer whose encoder self-attentions are replaced with an adaptation of Hyena for speech processing, where the long input sequences cause high computational costs.
Through experiments in automatic speech recognition (for English) and translation (from English into 8 target languages), we show that our best ConfHyena model significantly reduces the training time by 27\%, at the cost of minimal quality degradation ($\sim$1\%), which, in most cases, is not statistically significant.
\\
\newline }

\begin{document}

\maketitleabstract

\section{Introduction}
\label{sec:intro}
\blfootnote{\includegraphics[scale=0.1]{spotted_hyena96.png} The authors contributed equally.}
The attention mechanism \citep{attention} is the core of today's neural architectures in many 
AI fields \citep{lin2022survey}, including speech processing \citep{latif2023transformers}.
However, attention is known to be computationally expensive due to its quadratic computational and memory complexity with respect to the input length, which hinders the adoption of attention-based models in use cases that entail long input sequences.
This has complicated their application in speech tasks like automatic speech recognition (ASR) and speech translation (ST), where audio sequences are typically $\sim$8-10 times longer than the equivalent text.
In fact, current models have to downsample the input sequence by a factor of 4 with strided convolutions before applying attention-based layers \cite{berard_2018,di-gangi-etal-2019-enhancing}.
Besides causing information loss \citep{salesky-etal-2019-exploring-ph,papi-etal-2021-speechformer}, this workaround does not entirely solve the problem. The sequences can indeed remain very long, thereby resulting in high training costs with important social and environmental negative impact \citep{strubell-etal-2019-energy}.

While attention alternatives with sub-quadratic complexity have been proposed in other fields \citep{10.1145/3530811}, none of them have achieved widespread adoption, as their efficiency comes at the cost of non-negligible performance degradation in many tasks, including ASR and ST \citep{thesis_alastruey_2021}.
More recently, \citet{Poli2023HyenaHT} introduced the Hyena operator.
Based on a recurrence of implicitly parametrized long convolutions and data-controlled gating, Hyena has been the first attention alternative to claim comparable results (in language modeling and image classification).
Building upon these promising results, in this paper, we explore the adaptation and application of Hyena to ASR and ST, two tasks where training state-of-the-art models \citep{zhang2023google,pmlr-v202-radford23a} is extremely demanding in terms of computational resources, hardware, and time.\footnote{Attempts to build models similar to Whisper \citep{pmlr-v202-radford23a}, even on a smaller training dataset (less than 25\% of Whisper), require more than 15,000 GPU/hours on NVIDIA A100 GPUs \citep{peng2023reproducing}.}

Along this line of work, our contributions are:

\begin{enumerate}

\item{We adapt the original Hyena operator, which is \textit{causal} (i.e., it prevents accessing future information when encoding each element of a sequence), into a \textit{non-causal} version, allowing for richer representations by accessing information across the entire input sequence;}

\item{We present two models, ConfHyena and Hybrid ConfHyena,\footnote{\sara{Code and pre-trained models are released open-source under Apache 2.0 license at: \url{https://github.com/hlt-mt/FBK-fairseq}.}} based on the state-of-the-art Conformer \citep{gulati20_interspeech} architecture, which replace the self-attention with our adapted Hyena operator in \textit{all} encoder layers and \textit{only} in initial layers, respectively;}

\item{We show that on ASR (en) and ST (en$\rightarrow$\{de, es, fr, it, nl, pt, ro, ru\}), our Hybrid ConfHyena reduces training time by 27\%, at the cost of a minimal ($\sim$1\%) and mostly not statistically significant quality degradation.}
\end{enumerate}

\section{Background}

\subsection{Self-Attention}

Attention \sara{\citep{7472618}} is a function that maps a query matrix $Q$ and a pair of key-value matrices $K$ and $V$ to an output sequence $A(Q,K,V)$. 
In the case of self-attention, $Q$, $K$, and $V$ are computed from the same input sequence $x \in \mathbb{R}^{L \times d}$ through three different trainable matrices $W_q, W_k, W_v \in \mathbb{R}^{d \times d}$ as $Q=xW_q$, $K=xW_k$, and $V=xW_v$.
The scaled dot-product attention \sara{\citep{7472621}} used in the Transformer \citep{transformer} is then obtained by:

\begin{equation*}
   A(Q,K,V) = softmax \left( \frac{QK^T}{\sqrt{d_k}}  \right) V 
\end{equation*}
Since the matrix multiplication between $Q \in \mathbb{R}^{L \times d}$ and $K^T \in \mathbb{R}^{d \times L}$ results in a $\mathbb{R}^{L \times L}$ matrix, its overall memory and time complexity is $\mathcal{O}(L^2)$.

\subsection{\sara{Conformer}}
\label{subsec:conformer}

\mg{The self-attention is a cornerstone not only of the Transformer but also of the more recent Conformer \citep{gulati20_interspeech}, an architecture tailored for speech processing that significantly outperformed the Transformer in both ASR and ST \citep{inaguma2021non}. The Conformer modifies the encoder layer structure of the Transformer by 
\sara{implementing} three 
\sara{key modifications.}
First, it adds relative sinusoidal positional encodings \citep{dai-etal-2019-transformer} in\sara{to} the self-attention computation, eliminating the absolute positional embeddings added to the input in the Transformer. Second, it replaces the Transformer feed-forward network (FFN) with a Macaron-Net \citep{lu-et-al-2016-macaron-net}
\sara{consisting} of two FFNs that wrap the module. Lastly, it introduces a new convolution module after the self-attention. This module is made of a pointwise convolution, a Gated Linear Unit (GLU) activation function~\citep{Dauphin-2017-glu}, a depthwise convolution, a batch normalization \citep{ioffe-2015-batchnorm}, a Swish activation function \citep{swish-2017}, and another pointwise convolution.}

\subsection{Hyena}
\label{subsec:hyena}

Hyena \sara{\citep{Poli2023HyenaHT}} has been introduced as an alternative to self-attention with the goal of achieving the same representativeness (i.e., the ability to model dependencies between time steps of the input, regardless of their distance) while avoiding the quadratic complexity w.r.t. the input length.
The core of this operator is a recurrence of two operations: an implicitly parametrized long convolution (i.e., a convolution where the kernel has the same size as the input), and an element-wise multiplicative gating (i.e., an element-wise multiplication with a matrix of learned parameters). Hyena can be formalized as:

\begin{equation*}
   \begin{aligned}
 u_{0, ... , N}, z_0 & = \text{ShortConv}(W_{0, ..., N+1} x) \\
 z_{i+1} & = u_i \odot \text{LongConv}_{i}(z_{i}), i\in[0, N) \\
 y & = z_{N}
   \end{aligned}
\end{equation*}

\noindent
where $x$ is the input sequence, $W_{0, ..., N+1}$ are learned weights and $N$ is a hyperparameter named \textit{order}, which controls the depth of the recurrence and is usually set to 2.

The short convolution (ShortConv) models short-term dependencies and is a Conv1D with kernel size 3 and stride 1, which does not alter the shape of the input.
The long convolution (LongConv) models long-range dependencies and its efficient implementation is the key to the sub-quadratic complexity of the Hyena operator. 
A naive implementation of the LongConv would require learning a kernel of the same size as the input, which slides over the padded input $L$ times. This approach would result in a quadratic complexity, akin to self-attention.
In Hyena, instead, the kernel is implicitly learned by applying a
\mg{FFN}
over complex exponential positional embeddings \citep{Wang2020Encoding}, making the number of the kernel parameters 
independent from the input length.
Furthermore, Hyena bases its LongConv implementation on two principles:
\textit{i)} the convolution theorem \citep{10.1119/1.1973431}, which states that a circular convolution corresponds to an element-wise multiplication in the discrete Fourier domain \citep{10.5555/1795494}, and \textit{ii)} the equivalence between a linear convolution over a sequence of length $L$ with a kernel $k$ of length $K$ and a circular convolution with a kernel $k$ zero-padded to length $K+L$ \citep{oppenheim1975digital}. Building on them, Hyena computes the $\text{LongConv}_{i}(x)$ as: 

\begin{equation}
   \begin{aligned}
 z & = \text{iFFT}(\text{FFT}(\text{pad}(x)) \odot \text{FFT}(\text{pad}(k_i)) \\
 y & = z[:L]
 \label{eq:longconv}
   \end{aligned}
\end{equation}

\noindent
where FFT is the Fast Fourier Transform \citep{selesnick1997fast}, iFFT is the inverse FFT and $\text{pad}(.)$ pads the input with 0s up to a $2L$ length.
The output selection $z[:L]$ preserves \textit{causality} (i.e., the output at a given position depends only on the past), which is necessary for autoregressive models such as Transformer decoders.
The resulting LongConv and, in turn, Hyena implementation has a sub-quadratic computational complexity of $\mathcal{O}(L \log_2 L)$\mg{. While replacing the self-attention with sub-quadratic solutions based on the Fourier transform is not a new proposal \citep{lee-thorp-etal-2022-fnet}, Hyena has been the first to claim no performance loss}, which makes it a promising alternative to self-attention with quadratic complexity.

\begin{table*}[t]
\small
\setlength{\tabcolsep}{4pt}
   \centering
   \begin{tabular}{l||c|cccccccc|c}
       \hline
       \thickhline
       \textbf{Model} & \textbf{ASR (en)} & \textbf{en-de} & \textbf{en-es} & \textbf{en-fr} & \textbf{en-it} & \textbf{en-nl} & \textbf{en-pt} & \textbf{en-ro} & \textbf{en-ru} & \textbf{ST avg} \\
       \hline
       \hline
       Conformer & \textbf{10.64} & 24.97 & \textbf{30.48} & \textbf{36.43} & \textbf{26.25} & \textbf{30.31} & 30.09 & \textbf{24.67} & \textbf{17.35} & \textbf{27.57} \\
       \hline
       ConfHyena & 10.88 & 24.90 & 30.28 & 35.71* & 25.69* & 29.49* & 30.51* & 23.93* & 17.34 & 27.23 \\
       \quad - non-causal Hyena & 10.99* & 24.28 & 29.42* & 35.57* & 25.38* & 29.55* & 29.99 & 23.92* & 16.99* & 26.89 \\
       Hybrid ConfHyena & 10.75 & \textbf{25.22} & 30.15* & 36.19 & 26.04 & 29.76* & 30.43 & 23.78* & 17.06 & 27.33 \\
       \quad - non-causal Hyena & 11.27* & 24.84 & 30.28 & 36.09 & 25.81* & 29.53* & \textbf{30.69}* & 23.96* & 16.98* & 27.27 \\
       \thickhline
   \end{tabular}
   \caption{ConfHyena and Hybrid ConfHyena results on MuST-C v1.0 tst-COMMON for all language pairs. ASR scores are WER$\downarrow$ while ST scores are BLEU$\uparrow$. * means that the difference with the baseline (Conformer) is statistically significant as per bootstrap resampling \citep{koehn-2004-statistical} with 95\% CI.}
   \label{tab:main}
\end{table*}

\section{Hyena for Speech Processing}
\label{sec:speech_hyena}

As just seen, the original Hyena operator is built to preserve the causality property.
However, in the context of speech processing, constraining the encoder to access only past elements could reduce performance, as it is typically designed to look at the entire sequence to create context-aware encodings from the complete input representation \citep{chorowski2015attention}.
For this reason, we introduce \textbf{ConfHyena}, a Conformer-based model \sara{(\S\ref{subsec:conformer})} where 
encoder self-attentions are replaced by a non-causal version of the Hyena operator
\sara{that} can access the entire input sequence.
The differences with the causal version of Hyena are two:

\begin{enumerate}
   \item the ShortConv operator processes the current, preceding, and following time frames at each step, rather than the current and the two preceding frames.
   This is realized by setting the left padding to $\lceil \frac{K}{2} \rceil$ instead of $K-1$; 
   \item the LongConv operator is similarly adjusted, with the non-causal version implemented by altering the selection in Eq. \ref{eq:longconv} to $y = z[L/2:-L/2]$ to select the portion of the circular convolution that corresponds to a linear convolution with \textit{same} padding \citep{Goodfellow-et-al-2016}.
\end{enumerate}

With such modifications, ConfHyena has access to the same information as the self-attention in speech-processing models, while retaining a lower computational and memory complexity. 
The lower \textit{computational} complexity entails faster computation of long input sequences. 
The lower \textit{memory} complexity, instead, enables ConfHyena to process longer inputs compared to attention-based models without incurring out-of-memory issues and to run on GPUs with less VRAM.

In addition, we integrate a CTC-compression module \citep{gaido-etal-2021-ctc} into the middle of the encoder of all our systems.
This module \sara{is based on the connectionist temporal classification or CTC \citep{Graves2006ConnectionistTC} and} collapses intermediate context-aware encodings to reduce their redundancy, making them more effectively encoded by successive 
layers, with benefits in terms of both output quality and efficiency \citep{liu2020bridging,zhang-etal-2020-adaptive,zhao-etal-2022-redapt}.
As the sequences resulting from this module are significantly shorter, we posit that their length is sufficiently small so that they can be efficiently processed by self-attention.
As such, we propose \textbf{Hybrid ConfHyena}, an architecture where the non-causal Hyena operator is introduced only in the layers \sara{before} the CTC compression while preserving the self-attention in the subsequent ones.

\section{Experimental Settings}

For both ASR and ST, we use the MuST-C v1.0 dataset \citep{CATTONI2021101155}, comprising \textit{(audio, transcription, translation)} triplets in the TED-talks domain. 
Audios and transcriptions are in English while translations cover 8 languages (de, es, fr, it, nl, pt, ro, ru).
For ASR, we use the en-es section. 

The input is represented by 80-dimensional log mel-filterbank features (on 25ms windows sliding every 10ms) that are processed by 2 layers of CNN with stride 2, realizing a total downsampling of 4.
For all models, we use 12 encoder layers, 6 decoder layers, 8 heads, an embedding size of 512, and 2048 FFN width. 
The kernel size of both point- and depth-wise convolutions in Conformer layers is set to 31.
In Hyena operators, the order is set to 2, the width is 3 times the embedding size, and the filter FFN has 4 layers with 64 neurons and sine activations.
Dropout is set to 0.1 for all models.

\mg{To ensure the reliability of our findings, we used a Conformer implementation that is padding-safe (i.e., it does not change the output according to the amount of padding) and we tested our Hyena implementation with \texttt{pangolinn} to ensure the same property \citep{papi2023good}.}

We train with Adam optimizer, label-smoothed cross-entropy loss (smoothing factor 0.1), and CTC loss \citep{Graves2006ConnectionistTC} with weight factor 0.5 to ease convergence. We apply CTC compression after the 8\textsuperscript{th} encoder layer.
The learning rate is 2e-3 with Noam scheduler \cite{transformer} and 25,000 warm-up steps. SentencePiece unigram vocabularies \cite{kudo-richardson-2018-sentencepiece} are used with size 5,000 for English 
and 8,000 for all target languages.
We early stop the training after 10 epochs without improvement on the dev loss and average 5 checkpoints around the best.
We train on 2 NVIDIA A40 40GB VRAM GPUs with 40k tokens per batch and 4 as update frequency
and generate with 1 NVIDIA K80 16GB VRAM GPU. 
All other settings are the default of Fairseq-ST \cite{wang2020fairseqs2t}.
Evaluation is performed using WER for ASR and sacreBLEU \citep{post-2018-call}\footnote{BLEU|c:mixed|e:no|tok:13a|s:exp|v:2.0.0} for ST.

\section{Results}

For a comprehensive evaluation, in this section, we first analyze the output quality of ConfHyena and Hybrid ConfHyena (\S\ref{subsec:quality}), and then we study their efficiency (\S\ref{subsec:efficiency}), as both aspects are critical in determining the success of an architecture.

\subsection{Output Quality}
\label{subsec:quality}
Table \ref{tab:main} compares our ConfHyena and Hybrid ConfHyena models with Conformer in terms of transcription (ASR) and translation (ST) quality. 
Through ablation tests, we also report the results of the proposed architectures with the original causal Hyena operator (\textit{- non-causal Hyena}).

Focusing on the effect of the causality property, the results confirm its negative impact on model performance.
Indeed, the causal versions of both ConfHyena and Hybrid ConfHyena are consistently outperformed by those equipped with the non-causal operator both in ASR and ST.
The difference is more pronounced in ASR, where causality produces statistically significant degradations up to 0.52 WER.
This holds also for ST, albeit with less significant gaps (on average over the 8 language pairs, -0.34 and -0.06 BLEU for ConfHyena and Hybrid ConfHyena, respectively).
These results confirm the usefulness of accessing the entire input sequence to maximize 
transcription and translation quality and, therefore, \textbf{the superiority of the non-causal
Hyena} in speech encoders.

Comparing ConfHyena and Hybrid ConfHyena, the latter is superior both in terms of transcription and translation quality.
The gap between the two models, however, is marginal (-0.13 WER, +0.10 BLEU), showing that \textbf{the Hybrid variant has similar quality as the full ConfHyena encoder}.

Comparing Hybrid ConfHyena with Conformer, we observe that Conformer achieves the best results in most cases (ASR, and 6 out of 8 ST directions, with en-de and en-pt being the exceptions).
However, although Conformer consistently yields the best average score of 27.57 BLEU in ST, it is worth remarking that the overall margin over Hybrid ConfHyena is very narrow (-0.11 WER and +0.24 BLEU on average), corresponding to a $\sim$1\% relative difference. Moreover, the difference is statistically significant only in 3 ST directions.
We can therefore conclude that, in terms of mere output quality, \textbf{Hybrid ConfHyena performs closely to the attention-based Conformer model.}

\begin{table}[t]
\small
\setlength{\tabcolsep}{3pt}
   \centering
   \begin{tabular}{l||c|c|c}
       \hline
       \thickhline
       \multicolumn{1}{l}{\textbf{Model}} & \multicolumn{1}{c}{\textbf{\# param.}} &  \multicolumn{1}{c}{\textbf{train time$\downarrow$}} & \textbf{inf time$\downarrow$} \\
       \hline
       \hline
       Conformer & 114.9M & $\times$1.00 & $\times$1.00 \\
       \hline
       ConfHyena & \textbf{112.0M} &  $\times$1.04 & $\times$0.95 \\
       Hybrid ConfHyena & 112.9M & \textbf{$\times$0.73} & \textbf{$\times$0.93} \\
       \thickhline
   \end{tabular}
   \caption{Relative inference and training time averaged over all languages and tasks of Conformer, ConfHyena, and Hybrid ConfHyena.}
   \label{tab:time}
\end{table}

\subsection{Training and Inference Efficiency}
\label{subsec:efficiency}
After establishing that the output quality of the Conformer and ConfHyena models is similar, we now turn to assess their efficiency.
Looking at Table \ref{tab:time}, we notice that, in these terms, the differences are instead significant.

Notably, Hybrid ConfHyena emerges as the most efficient architecture by a large margin, reducing both training time by 27\% and inference time by 7\% in comparison to the Conformer model.
The greater savings in training time can be attributed to the autoregressive nature of the models.
In fact, while during training the number of forward (and backward) passes on the encoder and the decoder are the same (one per batch), during inference the encoder performs a single forward pass while most of the time is taken up by the multiple forward passes on the autoregressive decoder.

The training and inference time of ConfHyena is, instead, comparable to that of the Conformer, and much higher than that of Hybrid ConfHyena.
Although this may seem counterintuitive, the explanation is straightforward: the sub-quadratic complexity of Hyena makes it more efficient when sequences are long, while for shorter sequences, such as those obtained after the CTC compression, attention is faster.

In summary, we can conclude that \textbf{Hybrid ConfHyena substantially reduces computational costs compared to Conformer and ConfHyena}.
Drawing on this and the comparable output quality of the three models, we can state that \textbf{Hybrid ConfHyena offers the most favorable balance between quality and efficiency}.

\section{Reducing Downsampling}

The lower memory complexity of Hyena (as discussed in \S\ref{subsec:hyena}) opens up the possibility of reducing the initial downsampling performed by the two initial convolutions in speech encoders without incurring out-of-memory issues. 
A potential advantage of this operation would lie in mitigating the information loss caused by the context-uninformed compression of the input speech sequence. which is inherent to downsampling (see \S\ref{sec:intro}).
For this reason, we conclude this work by investigating the potential advantage of halving downsampling from a factor of 4 to a factor of 2.
In practice, this is obtained using stride 1 in the first convolution, while keeping stride 2 in the second.

The results in Table \ref{tab:analysis} show that there is no advantage in reducing the initial downsampling: in fact, halving it does not increase the translation ability of our models. Rather, this operation inflates both training and inference time by up to 50\% and 35\%, respectively. In light of this, we believe that future works should pursue other directions (e.g., different hyperparameters) to close the small quality gap without losing the efficiency gains.

\begin{table}[ht]
\small
\setlength{\tabcolsep}{3.5pt}
   \centering
   \begin{tabular}{l||c|c|c}
       \hline
       \thickhline
       \multicolumn{1}{l}{\textbf{Model}} & \multicolumn{1}{c}{\textbf{BLEU$\uparrow$}} & \multicolumn{1}{c}{\textbf{train time$\downarrow$}} & \textbf{inf time$\downarrow$} \\
       \hline
       \hline
       ConfHyena & 24.90 & \textbf{$\times$1.00} & \textbf{$\times$1.00} \\
       \quad + downsample 2 & 24.79 & $\times$1.39 & $\times$1.35 \\
       \hline
       Hybrid ConfHyena & \textbf{25.22} & \textbf{$\times$1.00} & \textbf{$\times$1.00} \\
       \quad + downsample 2 & \textbf{25.22} & $\times$1.50 & $\times$1.13 \\
       \thickhline
   \end{tabular}
   \caption{BLEU score and relative training/inference time of ConfHyena and Hybrid ConfHyena with downsample 2 on MuST-C en-de.}
   \label{tab:analysis}
\end{table}

\section{Conclusions}

In order to reduce the high computational costs of ASR and ST models, in this paper we proposed ConfHyena, a Confomer-based model that replaces self-attentions with non-causal Hyena operators, and a Hybrid version that mixes ConfHyena and Conformer layers. Our experiments on English ASR and 8 ST directions demonstrated that Hybrid ConfHyena has the best quality/efficiency trade-off, as it significantly reduces training time by 27\% with a minimal quality degradation of $\sim$1\% compared to the Conformer model.

\section*{Limitations}

\paragraph{The \enquote{good results} conundrum.}
By proposing and experimenting with Hybrid ConfHyena, we introduced an alternative architecture that improves model efficiency without substantial performance degradation. Admittedly, in our experiments, we recognize that Hybrid ConfHyena exhibits a gap, albeit limited, with respect to the current, state-of-the-art Conformer model. However, we refrain from interpreting our results as inherently \enquote{negative} or from considering a minor performance gap as an invalidating limitation. Indeed, we echo the recent criticism of purely leaderboard-based evaluation of new systems \citep{ethayarajh-jurafsky-2020-utility}, advocating for a more comprehensive perspective that considers a wider range of factors, including efficiency, ethics, and environmental sustainability \citep{sustainableai}. As such, we believe that achieving higher scores should not be the sole objective of research efforts, especially if it comes at the cost of resource-intensive training procedures \citep{ligozat2021unraveling} that sacrifice efficiency \citep{peng2023efficiency}. This approach also aligns with best practices recently adopted by major corporations that prioritize cost-effectiveness, environmental sustainability \citep{rolnick2022tackling}, and accessibility in model design \citep{companies4responsibleAI} \sara{to} reduce their impact and democratize their use.
Therefore, we underscore that the notable reduction in complexity and, in turn, in training time achieved by Hybrid ConfHyena should definitely be accounted as a positive advancement, regardless of the minimal quality degradation it implies.

\paragraph{Portability and scalability.} 
The validity of our findings should be confirmed across a wider range of datasets since our experiments only focused on the MuST-C corpus, which is the typical resource used in recent ST research.
The robustness of our results across different domains (MuST-C is composed of TED talks) and source languages (limited to English) have to be further verified, although there is no reason to believe that other settings may be more or less favorable to our proposed architecture. 
Furthermore, we did not scale our experiments to large training data, such as those employed in the training of state-of-the-art models like Whisper \citep{pmlr-v202-radford23a}, which are $\sim$100-1,000 times larger than MuST-C. 
Unfortunately, conducting such experiments is extremely expensive and demands access to high-performance hardware.
Evaluating the savings achievable by our proposed model in such a scenario is an interesting future step for this research, as well as confirming that the performance gap it suffers from remains limited.

\paragraph{\mg{Efficient Attention Implementation.}}
\mg{In our work we did not use efficient attention implementations\sara{,} such as Flash Attention \citep{dao2022flashattention}. While 
\sara{these} engineering optimizations can significantly reduce resource requirements and speed up computation, they are hardware-specific. For instance, the Flash Attention is not supported by the K80 GPUs used in our experimental settings. For this reason, we address the efficiency of the attention mechanism from a theoretical perspective, whose benefits are maintained regardless of the hardware employed.}

\section*{Acknowledgements}

We acknowledge the support of the PNRR project
FAIR - Future AI Research (PE00000013), under
the NRRP MUR program funded by the NextGenerationEU.

\section*{Bibliographical References}

\bibliographystyle{lrec-coling2024-natbib}
\bibliography{ref}

\end{document}